\documentclass[11pt]{article}

\usepackage{acl}

\usepackage{times}
\usepackage{latexsym}

\usepackage[T1]{fontenc}

\usepackage[utf8]{inputenc}

\usepackage{microtype}

\usepackage{inconsolata}

\usepackage{graphicx}

\title{From Anchors to Supervision: Memory-Graph Guided Corpus-Free Unlearning for Large Language Models}

\author{
  \textbf{Wenxuan Li\textsuperscript{1}},
  \textbf{Zhenfei Zhang\textsuperscript{1}},
  \textbf{Mi Zhang\textsuperscript{1}},
  \textbf{Geng Hong\textsuperscript{1}} \\
  \textbf{Mi Wen\textsuperscript{2}},
  \textbf{Xiaoyu You\textsuperscript{3}},
  \textbf{Min Yang\textsuperscript{1}} \\
  \textsuperscript{1}\small College of Computer Science and Artificial Intelligence, Fudan University, Shanghai, China \\
  \textsuperscript{2}\small Shanghai University of Electric Power, Shanghai, China \\
  \textsuperscript{3}\small School of Information Science and Engineering, East China University of Science and Technology, Shanghai, China \\
  \small \texttt{\{wxli24, zhangzf24\}@m.fudan.edu.cn}, \texttt{\{mi\_zhang, ghong, m\_yang\}@fudan.edu.cn}, \\
  \small \texttt{miwen@shiep.edu.cn}, \texttt{xiaoyuyou@ecust.edu.cn}
}

\usepackage{tikz}
\usepackage[most]{tcolorbox}
\usepackage{algorithm}
\usepackage{amsfonts}
\usepackage{amsmath}
\usepackage{amssymb}
\usepackage{amsthm}
\usepackage{bm}
\usepackage{comment}
\usepackage{caption}
\usepackage{framed}
\usepackage{mdframed}
\usepackage{listings}
\usepackage{makecell}

\usepackage{multirow}
\usepackage{booktabs} 
\usepackage{subcaption}
\usepackage[table]{xcolor}

\usepackage{pifont} 
\newcommand{\cmark}{\ding{51}} 
\newcommand{\xmark}{\ding{55}} 

\newcommand{\ourname}{MAGE}

\usepackage{xcolor}

\newmdenv{promptbox}
\mdfdefinestyle{promptbox}{
  linecolor=black,
  linewidth=1pt,
  roundcorner=10pt,
  font=\small
}

\usepackage{enumitem}
\setlist[itemize]{leftmargin=*}

\begin{document}
\maketitle

\begin{abstract}
Large language models (LLMs) may memorize sensitive or copyrighted content, raising significant privacy and legal concerns. While machine unlearning has emerged as a potential remedy, prevailing paradigms rely on user-provided forget sets, making unlearning requests difficult to audit and exposing systems to secondary leakage and malicious abuse.
We propose \textbf{\ourname}, a \textbf{M}emory-gr\textbf{A}ph \textbf{G}uided \textbf{E}rasure framework for \textbf{user-minimized, corpus-free} unlearning.
Given only a lightweight user anchor that identifies a target entity, \ourname\ probes the target LLM to recover target-related memorization, organizes it into a weighted local memory graph, and synthesizes scoped supervision for unlearning.
\ourname\ is model-agnostic and can be plugged into standard unlearning methods, and requires no access to the original training corpus.
Experiments on two benchmarks TOFU and RWKU demonstrate that \ourname\ 's self-generated supervision \textbf{achieves effective unlearning performance comparable to supervision generated with external reference}, while preserving overall utility.
These results support a practical and auditable unlearning workflow driven by minimal anchors rather than user-supplied forget corpora.

\end{abstract}

\section{Introduction}
\begin{figure}[t]
    \centering
    \includegraphics[width=\linewidth]{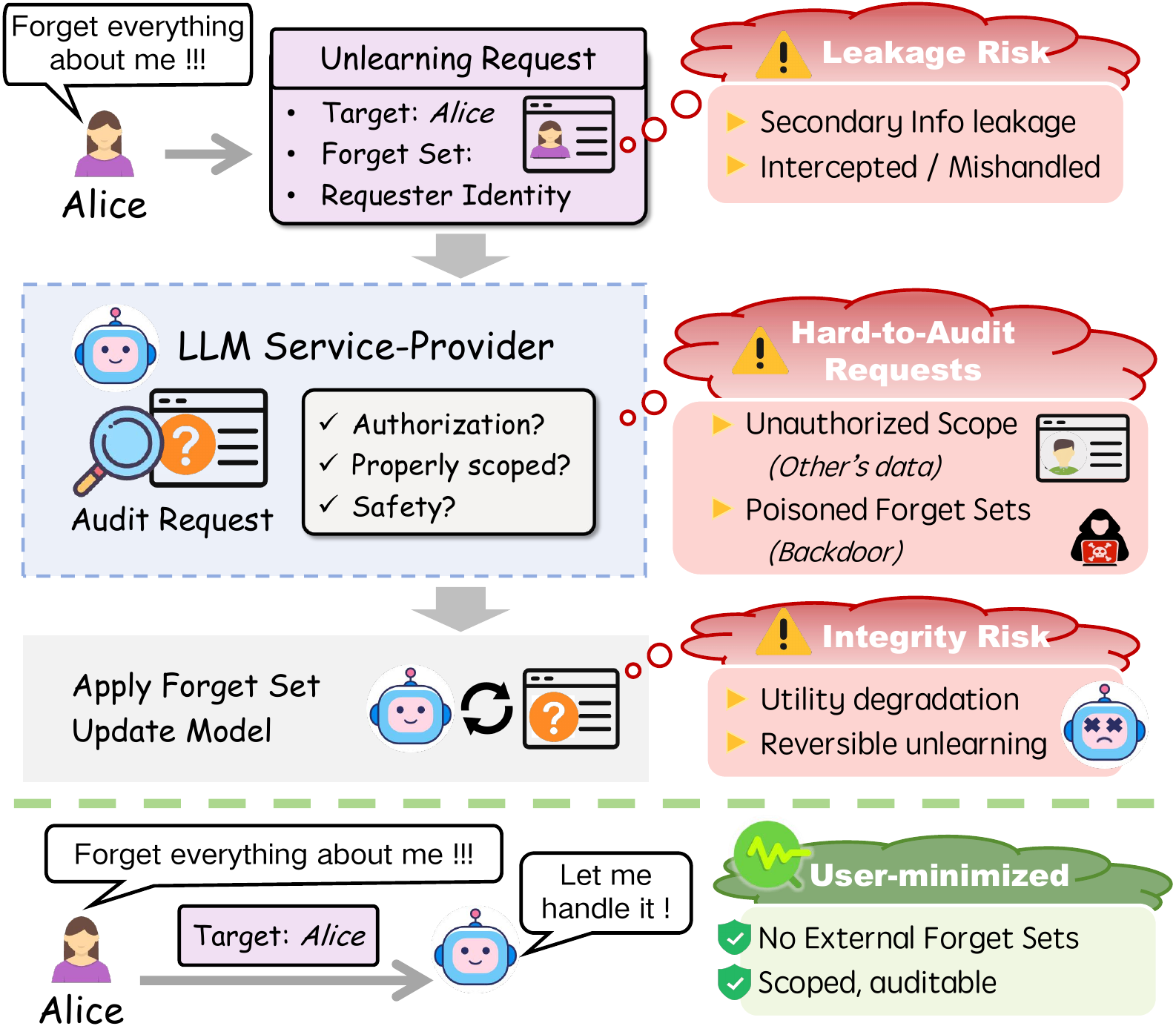}
    \caption{Risk surface of externally supplied forget sets in existing LLM unlearning paradigms. In contrast, a user-minimized paradigm mitigates these risks by improving information protection and auditability.}
    \label{fig:intro}
\end{figure}

Large language models (LLMs) have been widely deployed, yet their reliance on web-scale training data can lead to memorization of sensitive or copyrighted content \cite{carlini2021extracting, lucchi2024chatgpt}. A recent lawsuit by \textit{The New York Times} against OpenAI over unauthorized data use \cite{npr2025} further underscores these legal and ethical risks. Under regulations such as the General Data Protection Regulation (GDPR), individuals can invoke the “right to be forgotten” (RTBF) \cite{zhang2024right} to request removal of specific data from deployed models. 
To meet such requests without retraining from scratch, \textit{machine unlearning} has emerged as a practical approach to selectively erase targeted knowledge while preserving overall capabilities.

Most LLM unlearning pipelines fine-tune on a \texttt{forget set} that specifies the knowledge to erase, using objectives such as reverse optimization (e.g., GA \cite{jang2022knowledge}, NPO \cite{zhang2024negative}) or alternative-response training (e.g., WHP \cite{eldan2023s}, IDK \cite{ren2025general}). To preserve utility, a \texttt{retain set} of general-purpose data (e.g., Wikipedia) is typically included, and some settings further add a \texttt{neighbor set} consisting of distributionally similar but unlearning-irrelevant data to reduce collateral changes.
Despite recent progress, most existing approaches rely on a strong and often implicit assumption: the user requesting unlearning must provide a forget set that explicitly contains the information to be removed, which often contains sensitive content.

While this assumption underpins most existing unlearning pipelines, relying on an externally supplied forget set introduces practical risks that are often overlooked (Figure~\ref{fig:intro}).
From the user’s perspective, uploading sensitive content for unlearning can cause secondary information leakage~\cite{jin2024rwku}, especially if the forget corpus is intercepted or mishandled.
From the LLM service-provider’s perspective, externally submitted forget sets are inherently hard to audit, making unlearning requests vulnerable to abuse. 
Providers must verify authorization and scope, yet adversaries may submit wrongly scoped forget sets (e.g., containing other users’ data), leading to unauthorized forgetting. 
Moreover, seemingly benign forget sets can embed backdoors, which may stealthily degrade utility or make unlearning reversible~\cite{ren2025keeping,thaker2025position,shang2025forgetting}. 
Collectively, these leakage, auditability, and integrity risks cast doubt on the reliability of current \textit{externally provided forget set} paradigms .

To mitigate these risks, we advocate a more realistic corpus-free unlearning paradigm that minimizes user involvement to better match real-world deployment needs. Instead of submitting a full forget set, \textbf{the user provides minimal anchor information to specify the target}, making requests easier to audit and reducing privacy risks.
Despite these benefits, this paradigm introduces new challenges: (1) without an external forget set and with no access to the original training corpus during deployment, the LLM must reliably recover target-related memorization from its parameters to ensure effective unlearning; (2) the recovered memory must be translated into scoped supervision that separates what to forget from what to preserve.

We therefore take a step further and propose \textbf{\ourname}, a \textbf{m}emory-gr\textbf{a}ph \textbf{g}uided \textbf{e}rasure framework tailored to this user-minimized, corpus-free paradigm: it probes and reconstructs the LLM’s target-related memorization, abstracts it into a structured local memory graph, and uses the graph to self-generate scoped unlearning supervision.
For the first challenge, \ourname\ iteratively expands from minimal anchors to build a weighted local memory graph, and estimates memorization strength from the LLM’s outputs to filter noise and retain strongly memorized content.
For the second challenge, \ourname\ samples informative memory paths and converts them into a scoped forget set with a companion neighbor set, enabling focused unlearning while preserving model utility.

We validate our framework on two widely used entity unlearning benchmarks (TOFU \cite{maini2024tofu} and RWKU \cite{jin2024rwku}) across various unlearning strategies.
Our contributions are summarized as follows:
\begin{itemize}
    \item We identify key security risks in prevailing unlearning pipelines and introduce a \textbf{user-minimized, corpus-free} unlearning paradigm driven by minimal, auditable anchors.
    \item We develop \ourname, a memory-graph guided erasure framework that recovers target-related memorization and self-generates scoped unlearning supervision.
    \item Experiments on TOFU and RWKU show that \ourname\ achieves \mbox{supervised-level} unlearning performance using only self-generated data, supporting the feasibility of corpus-free unlearning.
\end{itemize}

\section{Related Work}
\subsection{LLM Unlearning}
LLM unlearning aims to remove specific knowledge from pre-trained LLMs, addressing privacy, security, and copyright concerns without full retraining \cite{liu2025rethinking}.
Most methods rely on parameter optimization to make the model behave as if it had never seen the forget set \cite{doshi2024does, lynch2024eight}. Some use gradient-ascent style updates to negate learned knowledge \cite{jang2022knowledge, yao2024large}, while others adopt preference-based objectives (e.g., DPO \cite{rafailov2023direct}, NPO \cite{zhang2024negative}, IDK \cite{ren2025general}, WHP \cite{eldan2023s}) to guide responses. To preserve utility, they typically incorporate a retain set or neighbor set via forward optimization, such as gradient descent \cite{liu2022continual}, KL regularization \cite{yuan2024closer}, or knowledge replacement \cite{xu2025relearn}. Recent work also improves unlearning performance by enhancing the quality of forget sets \cite{kuo2025proactive, wang2025selective, wang2025erasing}.
Beyond fine-tuning, researchers have explored other forgetting mechanisms, such as in-context learning \cite{pawelczyk2023context}, task vectors \cite{liu2024towards}, and data sanitization \cite{bhaila2024soft}. However, these indirect methods may still leave residual information. Knowledge editing enables fine-grained updates, but is less suitable for large-scale unlearning \cite{tian2024forget}.

\subsection{Entity-level unlearning}
Entity-level unlearning aims to remove a model’s knowledge about an entity, beyond the specific instances listed in a forget set. Since the target knowledge is not explicitly defined, it requires constructing representative forget data and evaluating broader effects on related knowledge.
TOFU \cite{maini2024tofu} introduced this setting with fictional entities, and later work \cite{ma-etal-2025-unveiling} showed that many methods still behave like instance-level unlearning and that performance is highly sensitive to forget-set quality. 
Recent studies improve entity-level unlearning through concept/parameter interventions \cite{wang2025erasing, choi-etal-2025-opt} or by augmenting forget data \cite{xu2025relearn}. 
However, most approaches assume predefined forget sets and largely overlook paradigm-level risks, whereas RWKU \cite{jin2024rwku} begins to highlight privacy concerns. We focus on entity-level unlearning because it aligns with real-world requests and challenges methods to remove entangled knowledge.

\subsection{Malicious Abuse of Unlearning Requests}
Recent work shows that unlearning requests become a security-critical interface when unlearning relies on user-supplied forget sets. In machine unlearning, carefully crafted requests can induce backdoor behaviors \cite{liu2024backdoor}, and auditing the submitted forget set alone may be insufficient to rule out stealthy attacks \cite{arazzi2025forgetting}. 
In the LLM setting, adversaries can associate common benign tokens with unlearning behavior to trigger broad utility degradation \cite{ren2025keeping} or make forgetting controllable and reversible \cite{shang2025forgetting}, while benchmark analyses suggest current evaluations may underestimate such risks \cite{thaker2025position}. 
Taken together, these findings both underscore the need to audit unlearning requests and expose how difficult it is to vet externally supplied forget corpora in practice, motivating settings driven by minimal, auditable anchors.
\section{Problem Definition}
In existing LLM unlearning settings, users requesting unlearning are required to provide an additional forget set $S_f$ containing the information to be erased. As illustrated in Figure~\ref{fig:intro}, this assumption can raise privacy risks and enables abuse. To address these concerns, we define a more practical user-minimized, corpus-free unlearning paradigm:

\paragraph{User Capability:} 
Users requesting unlearning for a target entity $E_t$ provide only a minimal anchor that identifies $E_t$, such as a name (e.g., ``\textit{Taylor Swift}'') or a short description to reduce ambiguity (e.g., ``\textit{Taylor Swift is a renowned American singer-songwriter.}''). 
They do not specify $S_f$ or the unlearning strategy, they only care whether the model can no longer reproduce or correctly reason about $E_t$, even under adversarial attempts to recover the forgotten information.

\paragraph{LLM Capability:}
The LLM (service provider) receives an unlearning request with a minimal anchor and no user-provided forget set. It must autonomously identify what to unlearn for a target entity $E_t$, define an appropriate unlearning scope, and apply an unlearning method $U$, while meeting the user’s request and limiting utility degradation.
Depending on deployment, it may self-construct a forget set $S_f$ (optionally with a neighbor set $S_n$) and use a retain set $S_r$ from a general-purpose distribution (e.g., Wikidata) to support scoped unlearning.
\section{Our Method}
\begin{figure*}[t]
    \centering
    \includegraphics[width=1\textwidth]{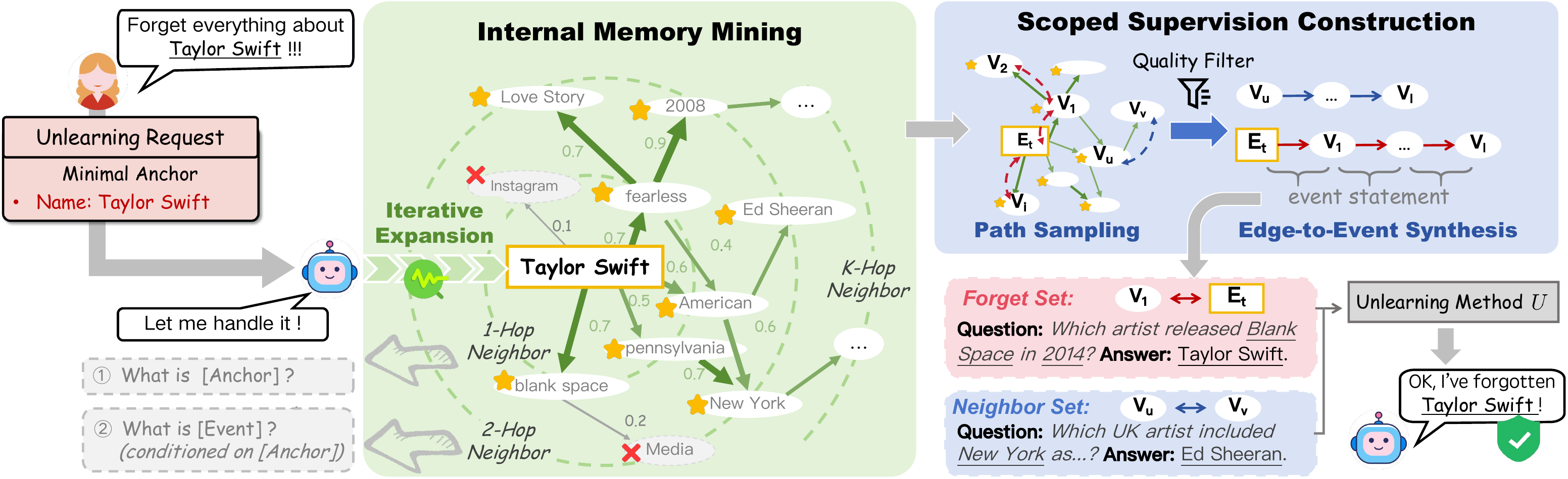}
    \caption{The framework of \ourname. \textbf{Internal Memory Mining}: Given an unlearning request, \ourname mines target memorization to build a strength-weighted local memory graph via iterative expansion. \textbf{Scoped Supervision Construction}: It then performs strength-weighted path sampling and edge-to-event synthesis to generate compact forget and neighbor supervision for downstream unlearning.}
    \label{fig:framework}
\end{figure*}
We propose \ourname, a memory-graph guided erasure framework with a two-stage pipeline: \textbf{Internal Memory Mining} reconstructs a weighted local memory graph around $E_t$, and \textbf{Scoped Supervision Construction} converts the graph into a scoped forget set $S_f$ with a companion neighbor set $S_n$ for downstream unlearning.

\subsection{Internal Memory Mining}
Directly prompting the model to describe $E_t$ often yields incomplete and unfocused recall. In corpus-free unlearning, missing memorized facts leads to ineffective forgetting, while noisy generations waste unlearning capacity.
Moreover, target knowledge in LLMs is often activated by related contextual cues \cite{hu2024jogging, patil2023can}, suggesting that effective recovery should iteratively expand from reliable anchors rather than rely on a single flat response.
We therefore iteratively expand from $E_t$ and represent the recovered memorization as a weighted \textit{local memory graph} $G_t=(V_t,\mathcal{E}_t)$, where $V_t$ contains salient target-related events, and $\mathcal{E}_t$ connects co-elicited items with weights reflecting memorization strength. This representation enables controllable expansion, explicit scope definition, and structured supervision synthesis.

\subsubsection{Self-Consistent Memorization}
We recover candidate $V_t$ by repeated elicitation. 
Concretely, we query the model $N$ times and apply entity extraction on each response to obtain candidate mentions.
We then score each candidate $v$ by mention frequency:
\begin{equation}
s(v)=\frac{1}{N}\sum_{i=1}^{N}\mathbb{I}\big[v \in \text{Entities}(y_i)\big],
\label{eq:freq_strength}
\end{equation}
where $y_i$ is the $i$-th elicited response.
We interpret $s(v)$ as a \textbf{memorization strength}: events repeatedly activated across independent generations are more likely to reflect stable memorized traces, while one-off mentions are downweighted as noise.

\subsubsection{Target-Conditioned Iterative Expansion}
We construct $G_t$ by iteratively expanding from $E_t$ up to $K$ hops.
At hop $0$, $V_t$ contains only the target node $E_t$.
We then elicit $N$ responses anchored at $E_t$, extract events, and score each candidate $v$ by its mention frequency $s(v)$ in Eq.~\ref{eq:freq_strength}. 
Candidates with $s(v)\ge\tau$ are retained as the 1-hop neighbors of $E_t$.
For hop $h$ ($h\!\ge\!1$), we further expand from the retained strong nodes $u$ in hop $h{-}1$.
To keep the expansion on-topic, we perform elicitation \emph{conditioned on $E_t$} with $u$ as a secondary anchor, extract new candidates, and again retain those with $s(\cdot)\ge\tau$.
We add directed edges $(u \rightarrow v)$ with weights reflecting co-elicitation salience:
\begin{equation}
w(u,v)=\frac{c(u,v)}{\sum_{v'\in \mathcal{N}(u)} c(u,v')},
\label{eq:edge_weight}
\end{equation}
where $c(u,v)$ counts how often $v$ is extracted from responses anchored at $u$, and $\mathcal{N}(u)$ denotes the extracted neighbors of $u$.
The resulting weighted graph $G_t$ summarizes what the model tends to recall around $E_t$ and which contextual cues most reliably trigger further recall.

Our elicitation prompts are provided in Appendix~\ref{apx:prompt_mining}, and we also discuss the overhead of memory-graph construction in Section~\ref{sec:overhead}.

\subsection{Scoped Supervision Construction}

Since $G_t$ is built based on \emph{co-elicitation}, we construct supervision from \textit{paths} rather than isolated nodes, which better reflects how target memories are triggered, enabling supervision that is naturally target-grounded and scoped.

\subsubsection{Weighted Path Sampling}
Prior work suggests that effective unlearning requires allocating stronger unlearning signals to content that is more strongly memorized, while weak or noisy traces contribute little useful supervision \cite{kuo2025proactive, tran2025tokens, wang2025selective, barbulescu2024each}. 
Motivated by this, we bias sampling toward high-strength regions of the memory graph, so that frequently sampled supervision aligns with the model’s memorization strength.

To focus supervision on strongly memorized content while maintaining coverage, we perform a strength-weighted random walk with exploration. Starting from $E_t$, the transition probability from $u$ to $v$ is:
\begin{equation}
p(v \mid u)\propto w(u,v)\cdot \Big(\frac{1}{1+\text{vis}(v)}\Big)^{\alpha},
\label{eq:walk_cov}
\end{equation}
where $\text{vis}(v)$ is the number of times $v$ has been visited in previous walks, and $\alpha$ controls exploration (larger $\alpha$ encourages coverage by downweighting frequently visited nodes). We collect $R$ walks with maximum length $L$ to obtain candidate paths.

We also apply a simple \textbf{path-quality} filter, discarding low-signal paths and retaining paths whose average edge weight is above a threshold:
\begin{equation}
q(\pi)=\frac{1}{|\pi|-1}\sum_{i=0}^{|\pi|-2} w(v_i,v_{i+1}) \ge \eta,
\label{eq:path_quality}
\end{equation}
where $\pi=(v_0,\ldots,v_{|\pi|-1})$ and $v_0=E_t$.
By sampling transitions in proportion to edge weights, the frequency with which nodes appear in sampled paths is roughly proportional to their memorization salience, so more strongly memorized content receives more unlearning supervision.

\subsubsection{Edge-to-Event Synthesis}

For each sampled path $(v_0{=}E_t, v_1, \dots, v_\ell)$, we use a sliding window to extract adjacent node pairs $(v_i, v_{i+1})$. Since each pair corresponds to a target-conditioned association mined during memory-graph construction, each pair provides a concrete contextual cue for targeted recall, rather than unconstrained and potentially irrelevant generation.
We prompt the LLM to produce a short, atomic \textbf{event statement} that links $v_i$ and $v_{i+1}$ in the context of $E_t$. We then set $\text{obj}=v_{i}$ and generate one concise forget sample (QA-style for illustration) grounded in the event statement: the question concerns $E_t$ and uses $\text{obj}$ as the primary context, while the answer uniquely identifies $E_t$.
We compute the unlearning loss only on the answer span, localizing gradient updates and reducing collateral changes. An example is shown below:
\begin{mdframed}[style=promptbox]
\texttt{Question: Which artist released \underline{Blank Space} in \underline{2014}? Answer: \textbf{Taylor Swift}.}
\end{mdframed}

To reduce collateral forgetting, we also construct a neighbor set $S_n$ that captures correlated but non-target information.
We sample paths starting from strong neighbors of $E_t$ while enforcing that $E_t$ does not appear on the sampled path, and generate corresponding QA samples whose answers are the non-target neighbor entities.
$S_n$ serves as a boundary constraint: it encourages the model to preserve nearby knowledge while forgetting target-specific content. An example is shown below:
\begin{mdframed}[style=promptbox]
\texttt{Question: Which UK artist included \underline{New York} as a hidden track on \underline{times} (2014)? \\Answer: \textbf{Ed Sheeran}}
\end{mdframed}
Our construction prompts are provided in Appendix~\ref{apx:prompt_corpus}.

Finally, we plug $(S_f, S_n)$ (and optionally a general retain set $S_r$) into an off-the-shelf unlearning method $U$. Since our output is standard fine-tuning data, \ourname\ is model-agnostic and compatible with existing unlearning pipelines.
\section{Experimental Settings}
\subsection{Evaluation Benchmark}
We evaluate \ourname\ on two representative entity-level unlearning benchmarks, which present two contrasting levels of memory-mining difficulty.

\paragraph{RWKU} \cite{jin2024rwku} targets real-world famous people. Since such knowledge is broadly present in pretrained LLMs, it is often possible to elicit abundant target-related facts and construct a rich memory graph. RWKU evaluates unlearning with four test groups: Forget Set ($\downarrow$), Neighbor Set ($\uparrow$), MIA Set, and Utility Set ($\uparrow$), where arrows indicate whether lower or higher is better (details in Appendix~\ref{apx:metric_rwku}).

\paragraph{TOFU} \cite{maini2024tofu} targets synthetic, fictitious authors generated by GPT-4 and introduced via benchmark-specific fine-tuning. Since these entities have limited supporting context and few co-occurrence signals in pretraining, the recovered memorization is often sparse and weakly connected, making it harder to mine a meaningful and coherent memory graph.
TOFU evaluates unlearning on a forget split and three utility splits (Retain, Real Authors, and World Facts), which form an increasing relevance gradient from in-domain to out-of-domain (Metric Details in appendix~\ref{apx:metric_tofu}). We run experiments on the \texttt{forget01} setting.

\begin{table}[t]
\centering
\caption{Priors and capabilities of forget-set generation methods. \cmark\ denotes used; \xmark\ denotes not used.}
\label{tab:baseline_priors}
\resizebox{\linewidth}{!}{
\begin{tabular}{lcccc}
\toprule
\textbf{Prior} & \textbf{RWKU} & \textbf{ELUDe} & \textbf{DirectQA} & \textbf{Ours} \\
\midrule
External Data (Wiki)         & \cmark & \cmark & \xmark & \xmark \\
Extra LLM / Human & \cmark & \cmark & \xmark & \xmark \\
Correctness Check & \cmark & \xmark & \cmark & \xmark \\
\bottomrule
\end{tabular}
}
\end{table}


\begin{table*}[htbp]
\centering
\resizebox{\linewidth}{!}{
\begin{tabular}{ll|cccc|ccc|cc|ccccc}
\toprule
Finetune & Method
& \multicolumn{4}{c|}{Forget Set $\downarrow$}
& \multicolumn{3}{c|}{Neighbor Set $\uparrow$}
& \multicolumn{2}{c|}{MIA Set}
& \multicolumn{5}{c}{Utility Set $\uparrow$} \\
 & 
& FB & QA & AA & All
& FB & QA & All
& FM $\uparrow$ & RM $\downarrow$
& MMLU & BBH & TruthfulQA & TriviaQA & AlpacaEval \\
\midrule
\multicolumn{2}{l|}{Before}
& 0.6788 & 0.6721 & 0.6829 & 0.6779
& 0.6306 & 0.7484 & 0.6895
& 1.9104 & 2.0433
& 0.4240 & 0.2654 & 0.2875 & 0.4056 & 6.2993 \\
\midrule
GA & RWKU
& \textbf{0.5817} & 0.5511 & 0.6536 & 0.5955
& 0.5600 & \textbf{0.7153} & 0.6376
& \textbf{3.1075} & 2.4128
& 0.3969 & 0.2346 & 0.3025 & \underline{0.3675} & 6.3109 \\
 & ELUDe
& 0.6440 & \underline{0.4646} & \underline{0.6026} & \underline{0.5704}
& 0.6135 & 0.6944 & \underline{0.6540}
& \underline{1.9933} & 2.1057
& \underline{0.4167} & \underline{0.2438} & \textbf{0.3175} & 0.3232 & \underline{6.3341} \\
 & DirectQA
& 0.6531 & 0.6675 & 0.7030 & 0.6745
& \textbf{0.6353} & \underline{0.6979} & \textbf{0.6666}
& 1.9286 & \textbf{2.0572}
& \textbf{0.4203} & 0.1975 & 0.2900 & \textbf{0.4640} & 6.2686 \\
\rowcolor{gray!15}
 & \ourname\
& \underline{0.6373} & \textbf{0.4288} & \textbf{0.5438} & \textbf{0.5366}
& \underline{0.6263} & 0.6501 & 0.6382
& 1.9885 & \underline{2.1016}
& 0.4130 & \textbf{0.2515} & \underline{0.3050} & 0.2955 & \textbf{6.3446} \\
\midrule
NPO & RWKU
& 0.6444 & 0.6128 & 0.6631 & 0.6401
& \underline{0.6103} & \underline{0.6925} & \underline{0.6514}
& \underline{2.1667} & 2.2093
& \underline{0.4137} & 0.2022 & 0.2925 & \underline{0.4053} & 6.3267 \\
 & ELUDe
& \underline{0.4290} & \underline{0.2718} & \underline{0.4061} & \underline{0.3690}
& 0.4829 & 0.4239 & 0.4534
& \textbf{2.1874} & 2.2383
& 0.3977 & 0.1867 & \textbf{0.3525} & 0.1526 & \textbf{6.3930} \\
 & DirectQA
& 0.6539 & 0.6842 & 0.7139 & 0.6840
& \textbf{0.6364} & \textbf{0.7213} & \textbf{0.6788}
& 1.9178 & \textbf{2.0494}
& \textbf{0.4211} & \underline{0.2099} & 0.2900 & \textbf{0.4658} & 6.2765 \\
\rowcolor{gray!15}
 & \ourname\
& \textbf{0.4177} & \textbf{0.1978} & \textbf{0.3734} & \textbf{0.3296}
& 0.4279 & 0.3800 & 0.4040
& 2.1054 & \underline{2.1770}
& 0.4123 & \textbf{0.2438} & \underline{0.3300} & 0.1656 & \underline{6.3849} \\
\midrule
GA+GD & RWKU
& \textbf{0.5108} & 0.4583 & 0.6075 & 0.5255
& 0.5691 & \textbf{0.7214} & \underline{0.6453}
& \textbf{7.1499} & 2.4194
& 0.3984 & 0.2191 & \underline{0.3050} & \underline{0.3708} & 6.2991 \\
 & ELUDe
& 0.5859 & \underline{0.4006} & \underline{0.5257} & \underline{0.5041}
& 0.5750 & 0.5864 & 0.5807
& 1.9600 & \underline{2.0716}
& 0.4152 & 0.1466 & \textbf{0.3350} & 0.2820 & \underline{6.3393} \\
 & DirectQA
& \underline{0.5468} & 0.6038 & 0.6772 & 0.6092
& \textbf{0.6464} & \underline{0.6953} & \textbf{0.6709}
& 1.8484 & \textbf{1.9758}
& \textbf{0.4174} & \textbf{0.2809} & 0.2775 & \textbf{0.4910} & 6.2556 \\
\rowcolor{gray!15}
 & \ourname\
& 0.5928 & \textbf{0.3644} & \textbf{0.5012} & \textbf{0.4861}
& \underline{0.5882} & 0.5788 & 0.5835
& \underline{2.0119} & 2.1121
& \underline{0.4159} & \underline{0.2741} & \underline{0.3050} & 0.2469 & \textbf{6.3499} \\
\midrule
GA+KL & RWKU
& \textbf{0.5108} & 0.4644 & 0.6128 & \textbf{0.5293}
& 0.5650 & \textbf{0.7222} & \underline{0.6436}
& \textbf{7.1544} & 2.4340
& 0.4050 & \underline{0.2160} & \underline{0.3050} & \underline{0.3718} & 6.3051 \\
 & ELUDe
& 0.6170 & \underline{0.4475} & \underline{0.5797} & 0.5480
& \underline{0.6148} & 0.6493 & 0.6320
& 1.9768 & \underline{2.0909}
& \underline{0.4167} & 0.1003 & \textbf{0.3175} & 0.3151 & \underline{6.3354} \\
 & DirectQA
& \underline{0.5812} & 0.6152 & 0.6968 & 0.6311
& \textbf{0.6239} & \underline{0.7064} & \textbf{0.6651}
& 1.9559 & \textbf{2.0635}
& 0.4152 & 0.2083 & 0.2850 & \textbf{0.4517} & 6.2774 \\
\rowcolor{gray!15}
 & \ourname\
& 0.6319 & \textbf{0.4364} & \textbf{0.5538} & \underline{0.5407}
& 0.6019 & 0.6295 & 0.6157
& \underline{2.0046} & 2.1068
& \textbf{0.4196} & \textbf{0.2358} & 0.3000 & 0.3291 & \textbf{6.3436} \\
\bottomrule
\end{tabular}}
\caption{Unlearning performance with different fine-tuning strategies and forget-set generation methods on Llama-2-7b-chat. Bold denotes the best result and underlined denotes the second best. Arrows indicate the direction of preference (\(\downarrow\) lower is better; \(\uparrow\) higher is better).}
\label{tab:rwku_res}
\end{table*}
\subsection{Forget-Set Generation Baselines}

Research on corpus-free forget-set construction remains limited. To contextualize \ourname, we compare against three representative forget-set generation baselines and summarize their reliance on external priors in Table~\ref{tab:baseline_priors}:\

\paragraph{RWKU} Uses Wikipedia passages for each target and GPT-4 to generate QA probes, then filters them with an open-source model and manual checks.

\paragraph{ELUDe}\cite{choi-etal-2025-opt} Uses the top-viewed Wikipedia pages as targets, generates paragraph-level QA pairs with GPT-4o, and deduplicates via Sentence-Transformer similarity.

\paragraph{DirectQA}\cite{ma-etal-2025-unveiling} Asks the target model to self-generate questions and answers about the entity, and keeps only self-verified QA pairs.

We evaluate \ourname\ with four fine-tuning strategies: GA, NPO, GA+GD, and GA+KL, detailed in appendix~\ref{apx:baseline_method}. Implementation details are provided in Appendix~\ref{apx:implement}.

\section{Experimental Results}
Our evaluation under the proposed corpus-free unlearning paradigm is designed to answer the following research questions. \textbf{RQ1}: Can LLM-recovered supervision match the forgetting performance of supervision built from ground-truth, externally provided data? \textbf{RQ2}: Does the recovered memory reflect what the LLM actually memorizes about the target entity? \textbf{RQ3}: What properties of the recovered data affect unlearning outcomes, including hallucinations, cross-model differences, and knowledge coverage?

\begin{table*}[htbp]
\centering
\scriptsize
\begin{tabular}{ll|cccc|cccc}
\toprule
Baseline & Generation Method
& \multicolumn{4}{c|}{Forget}
& \multicolumn{4}{c}{Utility} \\
 & 
& Prob$\downarrow$ & ROUGE$\downarrow$ & TruthRatio$\uparrow$ & Forget Q.$\uparrow$
& RS Score$\uparrow$ & RAS Score$\uparrow$ & WFS Score$\uparrow$ & Model Utility$\uparrow$ \\
\midrule
\multicolumn{2}{l|}{Before}
& 0.9951 & 0.9493 & 0.5306 & 0.0013
& 0.7133 & 0.1667 & 0.0508 & 0.6277 \\
\midrule
\multirow{3}{*}{GA}
& TOFU
& \textbf{0.0353} & \textbf{0.0714} & \textbf{0.5903} & \textbf{0.1650}
& 0.1906 & \textbf{0.1922} & \textbf{0.0725} & 0.3309 \\
& DirectQA
& 0.4260 & 0.4521 & 0.5029 & 0.0286
& \textbf{0.5378} & \underline{0.0830} & \underline{0.0212} & \textbf{0.5783} \\
\rowcolor{gray!15}
& \ourname\
& \underline{0.1492} & \underline{0.4500} & \underline{0.5285} & \underline{0.0541}
& \underline{0.4449} & 0.0046 & 0.0024 & \underline{0.5155} \\
\midrule
\multirow{3}{*}{NPO}
& TOFU
& \textbf{0.0359} & \textbf{0.0338} & \textbf{0.6272} & \textbf{0.1650}
& 0.1638 & \textbf{0.2402} & \textbf{0.0948} & 0.2951 \\
& DirectQA
& 0.4393 & 0.4933 & 0.5637 & 0.0541
& \textbf{0.5204} & \underline{0.0817} & \underline{0.0220} & \textbf{0.5743} \\
\rowcolor{gray!15}
& \ourname\
& \underline{0.0732} & \underline{0.2487} & \underline{0.5670} & \underline{0.0971}
& \underline{0.2613} & 0.0384 & 0.0046 & \underline{0.3949} \\
\bottomrule
\end{tabular}
\caption{Unlearning performance on TOFU-finetuned Llama-2-7b-chat. (↓ lower is better; ↑ higher is better)}
\label{tab:main_tofu}
\end{table*}

\subsection{Main Unlearning Results}

Table~\ref{tab:rwku_res} reports RWKU results under four fine-tuning strategies with different forget-set generation methods. For fair comparison with ELUDe, we average results over the eight overlapping targets shared by RWKU and ELUDe. 
Overall, \ourname\ is stable and competitive in the corpus-free setting, and it \textbf{reaches performance close to baselines that rely on externally provided supervision, supporting the feasibility of corpus-free unlearning (RQ1). }
Concretely, \ourname\ achieves the lowest Forget Set-All in three of four strategies (GA, NPO, GA+GD) and remains near the top under GA+KL, indicating strong forgetting effectiveness. While DirectQA occasionally yields higher Neighbor Set-All and better scores on a few utility metrics, \ourname\ consistently improves forgetting over DirectQA (lower Forget-All) with comparable utility retention, and it often attains the best AlpacaEval among corpus-free methods.

We further compare \ourname\ with the original TOFU supervision and the strongest corpus-free baseline, DirectQA in Table~\ref{tab:main_tofu}. 
Across both GA and NPO, \ourname\ produces a forget set that yields consistently stronger forgetting than DirectQA (lower Prob/ROUGE and higher Forget Q.), indicating that the mined memory graph provides more effective, target-specific supervision than direct recall prompting. Due to space limitations, GA+GD and GA+KL results are reported in appendix~\ref{apx:large_res_tofu}.
We additionally evaluate RWKU and TOFU on smaller-scale models (Phi-3-mini-4k and Llama-3.2-1B) in appendix~\ref{apx:small_res}.

\subsection{Memory Mining Analysis}\label{sec:memory_mining}
We analyze memory mining quality using two complementary protocols, as the availability of prior knowledge in LLMs differs across benchmarks.

\paragraph{Entity Similarity.}
For RWKU, which targets real-world public figures, target-related knowledge in pretrained LLMs is broad and entangled, and no ground-truth memorization is available. We therefore use ELUDe as a proxy reference and measure the similarity between entity memories mined by different data-generation methods.
\begin{figure}[t]
    \centering
    \includegraphics[width=\linewidth]{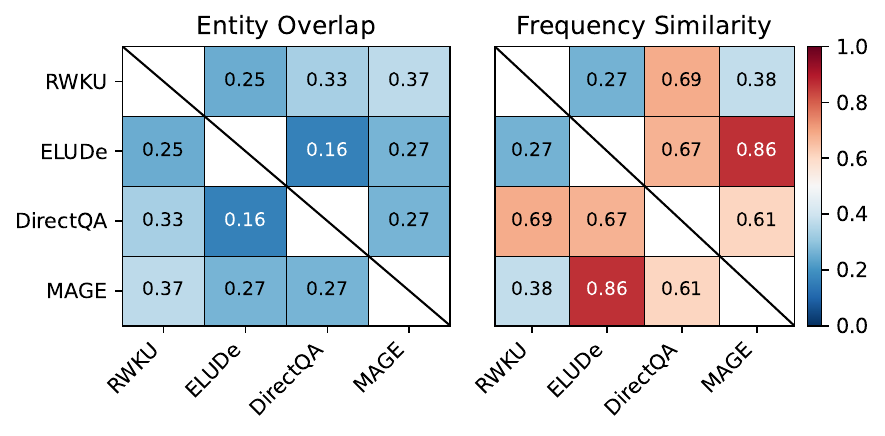}
\caption{Average entity overlap (Top-50 Jaccard) and frequency-distribution similarity across RWKU.}
    \label{fig:memory_rwku}
\end{figure}
Specifically, we extract entities from the forget set of each data-generation method and compare them from two perspectives: \textbf{Entity Overlap}, the Jaccard overlap of the top-50 entities ranked by frequency, and \textbf{Frequency Similarity}, the cosine similarity between normalized entity-frequency distributions after aligning entities to the union set. Figure~\ref{fig:memory_rwku} shows the similarity matrices averaged over all targets. Overall, entity overlap is low across methods, reflecting the entangled and diverse nature of real-world entity memories. In contrast, frequency similarity is notably higher, and \ourname\ achieves the strongest agreement with ELUDe (0.86), suggesting its mined entity salience best matches the proxy reference and providing evidence that our mined entity memories are closer to a plausible reference of pretrained memorization.

\paragraph{GT Attribute Recovery.}
\begin{table}[t]
\centering
\small
\begin{tabular}{l c c c c c}
\toprule
\multirow{2}{*}{\textbf{Attributes}} & \multirow{2}{*}{\textbf{Freq}} &
\multicolumn{2}{c}{\textbf{DirectQA}} & \multicolumn{2}{c}{\textbf{\ourname}} \\
\cmidrule(lr){3-4}\cmidrule(lr){5-6}
& & \textbf{Covered} & \textbf{F1} & \textbf{Covered} & \textbf{F1} \\
\midrule
BirthPlace & 2 & \ding{51} & 1.00 & \ding{51} & 0.80 \\
Gender     & 1 &           & 0    &           & 0    \\
BirthYear & 1 &           & 0    & \ding{51} & 0.67 \\
Genre      & 3 & \ding{51} & 0    & \ding{51} & 0.67 \\
Awards     & 1 &           & 0    & \ding{51} & 1.00 \\
Par\_Job   & 4 & \ding{51} & 0.80 &           & 0    \\
Books      & 3 &           & 0    & \ding{51} & 0.80 \\
Books      & 1 &           & 0    & \ding{51} & 0.33 \\
\bottomrule
\end{tabular}
\caption{Comparison of recovered attributes against TOFU ground-truth profiles.}
\label{tab:memory_tofu}
\end{table}
TOFU provides synthetic author profiles with defined ground-truth attributes, enabling a controlled evaluation of memory mining quality. We extract profile attributes from the generated forget set and compare them against the original TOFU profiles to quantify recovery accuracy; detailed extraction examples are deferred to the appendix~\ref{apx:tofu_memory_detail}).
As shown in Table~\ref{tab:memory_tofu}, DirectQA achieves limited coverage, suggesting that direct querying retrieves only a narrow subset of the injected profile information. In contrast, \ourname\ recovers a broader range of key attributes, highlighting the benefit of exploiting graph-structured memory signals. Nevertheless, \ourname\ still misses some non-entity-driven attributes, which may require targeted prompts or additional anchor information to elicit reliably.

Across RWKU and TOFU, the recovered memory from \ourname\ aligns well with plausible or ground-truth target knowledge, indicating that \textbf{it largely reflects what the LLM actually memorizes about the target entity (RQ2)}.

\paragraph{Memory Variation.}
Moreover, we observe notable cross-model differences in recovered memory graphs: for “Taylor Swift”, LLaMA-3-8b-instruct recalls “Taylor Alison Swift” as a close neighbor, while LLaMA-2-7b-chat does not, motivating model-specific forget supervision (Appendix~\ref{apx:memory_variation}).

\subsection{Forget Construction Analysis}
To answer \textbf{RQ3}, we examine how the properties of recovered supervision affect unlearning outcomes in the corpus-free paradigm. Unlike supervised pipelines with externally defined forget sets, corpus-free unlearning relies on model-recovered data whose quality can vary across targets and base models. Consequently, unlearning performance depends not only on the unlearning algorithm, but also on the recovered supervision itself.

\subsubsection{Hallucination}

\begin{figure}[t]
    \centering
    \includegraphics[width=\linewidth]{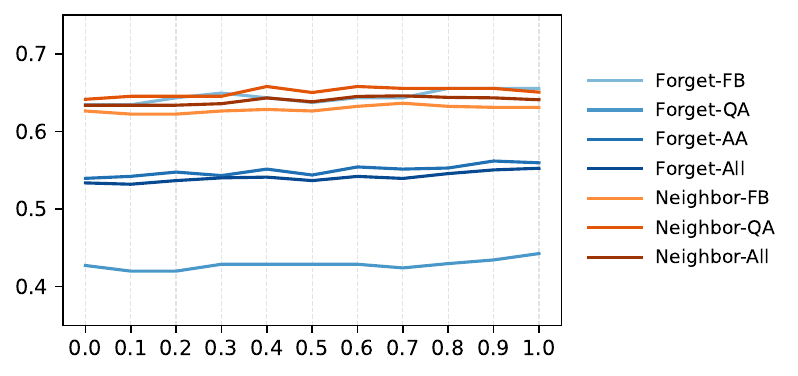}
\caption{Impact of the proportion of correct statements in the forget set on unlearning performance.}
    \label{fig:hallucination}
\end{figure}

Since corpus-free unlearning relies on LLM-generated supervision, the recovered forget set may include hallucinated QA pairs that sound plausible but are factually incorrect. To quantify the impact of such noise, we use GPT-4 to label each QA item in \ourname\ ’s RWKU forget set as correct or incorrect, then create mixed forget sets with correctness ratios from 0\% to 100\%. We run GA-based unlearning on each mixture and report the results in Figure~\ref{fig:hallucination}.
Surprisingly, performance changes only marginally across mixing ratios, suggesting that unlearning is not highly sensitive to the factual correctness of individual QA items. This further indicates that supervision aligned with the model’s memorized patterns may matter more than strict factual accuracy.
Additional results on smaller models, which typically exhibit more severe hallucinations, are provided in the appendix~\ref{apx:small_res}).

\subsubsection{Influence of the Neighbor Set}

We further study how the choice of the neighbor set affects unlearning performance. In Table~\ref{tab:rwku_res}, \textsc{GA+GD} and \textsc{GA+KL} use our recovered forget set together with our constructed neighbor set. We then replace the neighbor set with the Wiki data used by the baseline. 
As shown in Table~\ref{tab:neighbor_rwku}, this replacement consistently degrades performance on the neighbor set, indicating that the distribution of the neighbor data plays an important role in preserving unlearning performance.

\subsubsection{Knowledge Coverage}
In Table~\ref{tab:rwku_res}, we impose a high forget-set coverage threshold (e.g., $\ge$90\%) to ensure broad supervision over the target entity and enable thorough unlearning. We then vary this requirement and report the results in Figure~\ref{fig:coverage_ratio} (Appendix). As the threshold is relaxed, reduced coverage leads to weaker unlearning, while neighbor performance remains relatively stable. This aligns with prior findings that insufficient forget-set coverage is a key bottleneck for entity-level unlearning~\cite{ma-etal-2025-unveiling}.
\begin{table}[t]
\centering
\small
\begin{tabular}{llcc}
\toprule
 &  & \textbf{Forget-Avg} & \textbf{Neighbor-Avg} \\
\midrule
\multirow{2}{*}{\textbf{GA+GD}} & Wiki     & 0.4834 & 0.5644 \\
                              & Neighbor & \textbf{0.4861} & \textbf{0.5835} \\
\midrule
\multirow{2}{*}{\textbf{GA+KL}} & Wiki     & \textbf{0.5353} & 0.6030 \\
                              & Neighbor & 0.5407 & \textbf{0.6157} \\
\bottomrule
\end{tabular}
\caption{Impact of the Neighbor Set on Unlearning Performance}
\label{tab:neighbor_rwku}
\end{table}

\subsubsection{Recovery Overhead}\label{sec:overhead}
\ourname\ introduces a one-time memory-graph construction step, which is training-free and consists only of offline LLM interactions. We measure its cost by the number of elicitation iterations $I$, which is typically comparable to the number of recovered nodes $|V_t|$. If each iteration repeats a sub-query up to $N$ times, then $Q_{\text{LLM}} \le N\cdot I_{\max}$ and $\mathbb{E}[Q_{\text{LLM}}]\approx N\cdot I_{mean}$. Empirically, RWKU requires 88.36 iterations per entity on average, while TOFU needs only 13.75, indicating modest overhead in knowledge-sparse, user-facing cases. The recovered graph also provides a budgeted stopping criterion and can be reused across unlearning methods and repeated evaluations, amortizing its cost while improving coverage over DirectQA-style probing. In contrast, many alternatives incur hidden overheads such as external data collection/curation, retrieval and indexing, or additional filtering and judging calls, which \ourname\ avoids by relying on bounded model interactions.
\section{Conclusion}

We presented \ourname, a memory-graph guided erasure framework for user-minimized, corpus-free LLM unlearning. Motivated by the auditability and security risks of user-supplied forget sets, \ourname\ recovers target-related memorization, organizes it into a weighted local memory graph, and self-generates scoped supervision that plugs into standard unlearning methods without access to the training corpus. Experiments on TOFU and RWKU show that \ourname\ achieves supervised-level forgetting with comparable utility retention using only self-generated data, supporting a practical and auditable workflow driven by minimal anchors.

\bibliography{anthology}

\clearpage
\appendix
\appendix

\section{Experiment Settings} 
\subsection{Implementation Details} \label{apx:implement}

\noindent\textbf{Models.}
We run RWKU experiments on the base pretrained models \texttt{Llama-2-7b-chat} \cite{touvron2023llama} and \texttt{Phi-3-mini-4k-instruct} \cite{abdin2024phi3}. 
For TOFU, we use benchmark-specific fine-tuned checkpoints released by prior work on \texttt{Llama-2-7b-chat}\footnote{\url{https://huggingface.co/locuslab/tofu_ft_llama2-7b}} and \texttt{Llama-3.2-1B-Instruct}\footnote{\url{https://huggingface.co/open-unlearning/tofu_Llama-3.2-1B-Instruct_full}} since TOFU requires the injected author profiles to be present in the model.

\noindent\textbf{Fine-tuning.}
All unlearning pipelines are implemented with LoRA on NVIDIA GeForce RTX 4090 GPU. 
We use LoRA rank $r{=}8$ (RWKU) and $r{=}16$ (TOFU), with $\alpha{=}16$.
To enable fair comparisons across different forget-set generation methods under the same unlearning strategy, we fix the learning rate within each strategy and adjust the number of epochs to keep the total number of optimization steps approximately matched, accounting for differences in dataset size. 
For RWKU, we largely follow ELUDe’s tuning protocol but slightly adjust learning rates for LoRA stability: GA uses $8\!\times\!10^{-6}$ for 4 epochs, NPO uses $5\!\times\!10^{-5}$ for 4 epochs, and GA+GD / GA+KL use $5\!\times\!10^{-6}$ for 3 epochs, while epochs for other forget-set sources are scaled to match steps. 
For TOFU, we fix the learning rate to $2\!\times\!10^{-4}$ across all settings and tune epochs to keep optimization steps comparable; the TOFU benchmark supervision uses 4 (GA), 6 (NPO), 3 (GA+GD), and 6 (GA+KL) epochs, and we adjust DirectQA and \ourname\ accordingly.

\subsection{Method Hyperparameters} \label{apx:method_hparams}
\noindent\textbf{Internal Memory Mining.}
We build the local memory graph by iterative, target-conditioned elicitation and frequency-based filtering.
We use $N{=}10$ elicitation queries per anchor, keep a candidate node if its memorization-strength score satisfies $s(\cdot)\ge\tau$, with $\tau{=}0.2$ on RWKU and $\tau{=}0.3$ on TOFU.
We expand the graph up to $K$ hops, with $K{=}2$ on RWKU and $K{=}3$ on TOFU.

In addition to a fixed $K$, we observe that as expansion proceeds, outer-hop neighbors tend to have fewer strong connections back to $E_t$; this suggests an adaptive stopping rule that halts expansion when the aggregate strength of edges pointing back to $E_t$ drops below a threshold.

\noindent\textbf{Scoped Supervision Construction.}
We set the exploration coefficient to $\alpha{=}1$, the maximum walk length to $L{=}5$, and the path-quality threshold to $\eta{=}0.3$. We run $R{=}200$ walks for RWKU to match the supervision scale of ELUDe, and $R{=}20$ walks for TOFU due to its sparser recovered memory.

\subsection{Evaluation Metrics}\label{apx:metric}

\subsubsection{RWKU Evaluation Metrics}\label{apx:metric_rwku}
RWKU \cite{jin2024rwku} reports results in four groups: Forget Set ($\downarrow$), Neighbor Set ($\uparrow$), MIA Set, and Utility Set ($\uparrow$).

\begin{table*}[htbp]
\centering
\resizebox{\linewidth}{!}{
\begin{tabular}{ll|cccc|ccc|cc|ccccc}
\toprule
Finetune & Method
& \multicolumn{4}{c|}{Forget Set $\downarrow$}
& \multicolumn{3}{c|}{Neighbor Set $\uparrow$}
& \multicolumn{2}{c|}{MIA Set}
& \multicolumn{5}{c}{Utility Set $\uparrow$} \\
 & 
& FB & QA & AA & All
& FB & QA & All
& FM $\uparrow$ & RM $\downarrow$
& MMLU & BBH & TruthfulQA & TriviaQA & AlpacaEval \\
\midrule
\multicolumn{2}{l|}{Before}
& 0.6729 & 0.7290 & 0.6839 & 0.6953
& 0.6251 & 0.5819 & 0.6035
& 1.8448 & 1.9326
& 0.6864 & 0.4090 & 0.3775 & 0.3985 & 6.0196 \\
\midrule
GA & RWKU
& \underline{0.6597} & 0.7404 & 0.6804 & 0.6935
& \underline{0.6227} & \underline{0.5692} & \textbf{0.5959}
& 1.8645 & \underline{1.9411}
& \textbf{0.6879} & 0.3133 & 0.3800 & \textbf{0.4077} & 6.0383 \\
 & ELUDe
& 0.6694 & 0.6799 & 0.6625 & 0.6706
& 0.6113 & \textbf{0.5793} & 0.5953
& 1.8567 & \textbf{1.9383}
& \underline{0.6857} & 0.3148 & \underline{0.3925} & 0.2670 & \textbf{6.1704} \\
 & DirectQA
& 0.6675 & \textbf{0.6541} & \textbf{0.6296} & \underline{0.6504}
& \textbf{0.6377} & 0.5534 & \underline{0.5956}
& \textbf{2.0591} & 2.0101
& 0.6784 & \underline{0.4336} & 0.3775 & \underline{0.3506} & 6.0679 \\
\rowcolor{gray!15}
 & \ourname\
& \textbf{0.6236} & \underline{0.6724} & \underline{0.6355} & \textbf{0.6438}
& 0.6083 & 0.5590 & 0.5837
& \underline{1.8663} & 1.9439
& 0.6842 & \textbf{0.4383} & \textbf{0.3975} & 0.2460 & \underline{6.1535} \\
\midrule
NPO & RWKU
& \textbf{0.6245} & 0.7129 & 0.6561 & 0.6645
& \underline{0.6300} & \textbf{0.5688} & \textbf{0.5994}
& \textbf{2.1448} & 2.1175
& 0.6827 & 0.4012 & 0.3900 & \underline{0.2253} & 6.2853 \\
 & ELUDe
& 0.6793 & \textbf{0.5073} & \underline{0.4865} & \underline{0.5577}
& 0.5971 & 0.4836 & 0.5404
& 1.9096 & \underline{1.9610}
& \underline{0.6842} & 0.3858 & \textbf{0.4475} & 0.1546 & \textbf{6.4073} \\
 & DirectQA
& 0.6497 & 0.6870 & 0.6826 & 0.6731
& 0.6183 & \underline{0.5561} & 0.5872
& \underline{1.9156} & 1.9630
& \textbf{0.6849} & \underline{0.4213} & 0.3875 & \textbf{0.3771} & 6.0951 \\
\rowcolor{gray!15}
 & \ourname\
& \underline{0.6283} & \underline{0.5097} & \textbf{0.4550} & \textbf{0.5310}
& \textbf{0.6348} & 0.5491 & \underline{0.5920}
& 1.8984 & \textbf{1.9548}
& 0.6835 & \textbf{0.4290} & \underline{0.4325} & 0.1642 & \underline{6.3429} \\
\midrule
GA+GD & RWKU
& \underline{0.6356} & 0.7296 & 0.6675 & 0.6776
& 0.5952 & \textbf{0.5772} & \textbf{0.5862}
& \textbf{1.8965} & 1.9507
& \underline{0.6871} & \underline{0.3441} & 0.3750 & 0.4245 & \underline{6.0307} \\
 & ELUDe
& \underline{0.6356} & \textbf{0.6163} & \textbf{0.6440} & \textbf{0.6386}
& \textbf{0.6084} & 0.5624 & \underline{0.5854}
& 1.8513 & \textbf{1.9342}
& \textbf{0.6893} & 0.2238 & \textbf{0.3975} & 0.2859 & \textbf{6.1453} \\
 & DirectQA
& 0.6422 & 0.7201 & 0.6545 & 0.6723
& 0.5827 & \underline{0.5702} & 0.5765
& \underline{1.8635} & 1.9436
& 0.6820 & \underline{0.3441} & \underline{0.3825} & \textbf{0.4515} & 5.8489 \\
\rowcolor{gray!15}
 & \ourname\
& 0.6484 & \underline{0.7042} & \underline{0.6515} & \underline{0.6680}
& \underline{0.5995} & 0.5646 & 0.5821
& 1.8592 & \underline{1.9413}
& 0.6857 & \textbf{0.4306} & \underline{0.3825} & \underline{0.4408} & 5.9923 \\
\midrule
GA+KL & RWKU
& \textbf{0.6293} & 0.7296 & \underline{0.6684} & 0.6758
& \underline{0.6131} & \textbf{0.5746} & \underline{0.5938}
& \textbf{1.8951} & 1.9506
& 0.6849 & 0.3997 & 0.3750 & \textbf{0.4113} & 6.0506 \\
 & ELUDe
& 0.6828 & \underline{0.6847} & 0.6722 & 0.6799
& 0.6072 & \underline{0.5710} & 0.5891
& 1.8559 & \underline{1.9376}
& \textbf{0.6879} & \underline{0.4090} & \textbf{0.3950} & 0.2730 & \textbf{6.1676} \\
 & DirectQA
& 0.6608 & \textbf{0.6529} & \textbf{0.6535} & \textbf{0.6557}
& \textbf{0.6685} & 0.5491 & \textbf{0.6088}
& \underline{1.8807} & 1.9429
& 0.6857 & 0.3426 & \underline{0.3850} & \underline{0.3752} & 6.0251 \\
\rowcolor{gray!15}
 & \ourname\
& \underline{0.6330} & 0.7135 & 0.6694 & \underline{0.6720}
& 0.5808 & 0.5691 & 0.5749
& 1.8535 & \textbf{1.9370}
& \underline{0.6864} & \textbf{0.4306} & \underline{0.3850} & 0.2969 & \underline{6.1099} \\
\bottomrule
\end{tabular}}
\caption{Unlearning performance with different fine-tuning strategies and forget-set generation methods on Phi-3-mini-4k-instruct. Bold denotes the best result and underlined denotes the second best. Arrows indicate the direction of preference (\(\downarrow\) lower is better; \(\uparrow\) higher is better).}
\label{tab:apx_rwku_res}
\end{table*}

\paragraph{Forget Set ($\downarrow$).}
RWKU measures whether a model still reveals target knowledge using three probe types:
(i) FB (fill-in-the-blank) probes that mask knowledge points in sentences from the target's Wikipedia page and ask the model to complete the blank;
(ii) QA (question-answer) probes that directly query target-related facts; and
(iii) AA (adversarial-attack) probes that attempt to elicit forgotten knowledge via prompt attacks (e.g., prefix injection, affirmative suffix, role playing, reverse query, synonym manipulation, background hints, in-context demonstrations, and cross-lingual queries).
For FB/QA/AA, RWKU uses ROUGE-L recall between the model output and the ground-truth answer; lower is better for unlearning efficacy. We additionally report All as the aggregate over the three forget subsets, consistent with RWKU’s reporting format.

\paragraph{Neighbor Set ($\uparrow$).}
To assess locality (i.e., not over-forgetting beyond the target), RWKU constructs neighbor probes that are closely related to the target but outside the intended forgetting scope. Similar to the forget set, it includes Neighbor FB and Neighbor QA, evaluated by ROUGE-L recall; here higher is better, indicating less collateral damage on neighboring knowledge. We report All as the aggregate over the neighbor subsets.

\paragraph{MIA Set (FM $\uparrow$, RM $\downarrow$).}
RWKU further evaluates whether target knowledge still appears as pretraining members via membership inference attacks (MIAs). It defines a forget-member set (FM) containing target-related training fragments and a retain-member set (RM) with unrelated member fragments. RWKU provides four MIA scorers (LOSS, Zlib entropy, Min-K\% Prob, and Min-K\%++ Prob) and primarily reports LOSS-based scores; higher scores indicate the text is less likely to be memorized. Therefore, effective unlearning should yield higher FM (target fragments look non-member) while keeping lower RM (unrelated member fragments still look like members), matching the (FM $\uparrow$, RM $\downarrow$) convention.

\paragraph{Utility Set ($\uparrow$).}
RWKU measures general capability preservation with five standard tasks, where higher is better:
MMLU (Gen) using 5-shot accuracy via answer perplexity,
BBH (Rea) using chain-of-thought prompting with 3-shot examples and exact-match scoring,
TruthfulQA (Tru) using MC1 6-shot accuracy,
TriviaQA (Fac) using 6-shot F1, and
AlpacaEval (Flu) using the weighted average of bi-/tri-gram entropies as a fluency proxy.

\subsubsection{TOFU Evaluation Metrics}\label{apx:metric_tofu}
TOFU \cite{maini2024tofu} evaluates unlearning with four splits: Forget Set (target to forget), and three non-forget splits for utility preservation—Retain Set, Real Authors, and World Facts—forming a relevance gradient from near-domain to far-domain \cite{maini2024tofu}.

\paragraph{Forget metrics.}
On the Forget Set, we report: (i) Prob$\downarrow$, the length-normalized conditional likelihood of the ground-truth answer under the model, computed as $P(a\mid q)^{1/|a|}$ (lower is better when forgetting) \cite{maini2024tofu}; (ii) ROUGE$\downarrow$, the ROUGE-L recall between the greedy-decoded answer and the ground-truth answer (lower indicates less recoverable target content) \cite{maini2024tofu}; (iii) TruthRatio$\uparrow$, a likelihood-ratio style score that contrasts a paraphrased correct answer $\tilde{a}$ against a set of GPT-generated perturbed (factually incorrect) answers $\mathcal{A}_{\mathrm{pert}}$, defined as
$\log R_{\mathrm{truth}}=\frac{1}{|\mathcal{A}_{\mathrm{pert}}|}\sum_{\hat{a}\in\mathcal{A}_{\mathrm{pert}}}\frac{1}{|\hat{a}|}\log P(\hat{a}\mid q)-\frac{1}{|\tilde{a}|}\log P(\tilde{a}\mid q)$,
where $\tilde{a}$ and $\mathcal{A}_{\mathrm{pert}}$ are constructed to control for phrasing effects \cite{maini2024tofu}; and (iv) Forget Q.$\uparrow$, the forget quality score computed via a two-sample KS-test on the TruthRatio distributions of the unlearned model versus a retain-only model, using the resulting $p$-value as the forgetting score (higher $p$ implies the two models are harder to distinguish, i.e., better forgetting) \cite{maini2024tofu}. Consistent with the TOFU convention, Forget Set uses ``lower Prob/ROUGE, higher TruthRatio'' as the desired direction \cite{maini2024tofu}.

\paragraph{Utility metrics.}
For utility preservation, TOFU rescales metrics so that higher is better on non-forget splits (e.g., converting TruthRatio by $\max(0,1-R_{\mathrm{truth}})$ on non-forget sets) and aggregates them into Model Utility$\uparrow$ by taking the harmonic mean over the three metrics (Prob/ROUGE/TruthRatio) across the three non-forget splits (Retain, Real Authors, World Facts), yielding nine values in total; the harmonic mean penalizes any single metric collapse \cite{maini2024tofu}. 
In addition, we report per-split summary scores in the same order as the table header: RS Score$\uparrow$ (Retain Set), RAS Score$\uparrow$ (Real Authors Set), and WFS Score$\uparrow$ (World Facts Set). Each score is computed on its corresponding split as the harmonic mean of Prob., ROUGE, and an accuracy-oriented term (Acc.), so that higher indicates better overall utility on that split.

\subsection{Basic Fine-tuning Strategies} \label{apx:baseline_method}
\ourname\ is a frame work can be integrated into existing unlearning pipelines, we choose the following:
\paragraph{Gradient Ascent (GA)} inverts the standard training objective for forget set $S_f$ by ascending the gradient of the loss function:
\begin{equation*}
    \theta \leftarrow \theta + \eta \nabla_{\theta} \mathcal{L}_{\text{LM}}(S_f)
\end{equation*}
Here, $\theta$ represents the model's parameters, $\eta$ is the unlearning rate (a hyperparameter that controls the step size), and $\nabla_{\theta} \mathcal{L}_{\text{LM}}(S_f)$ is the gradient of the language model loss calculated on the forget set $S_f$.

\paragraph{Negative Preference Optimization (NPO)} trains the model to treat the generation of forgotten content as a rejected behavior, contrasting it with the behavior of a reference model:
\begin{equation*}
\mathcal{L}_{\text{NPO}} = - \mathbb{E}_{x \sim S_f} \left[ \log \sigma \left( \beta \log \frac{\pi_{\theta_{\text{ref}}}(y|x)}{\pi_{\theta}(y|x)} \right) \right]
\end{equation*}
Here, $\pi_{\theta}$ is the policy of the model being trained, $\pi_{\theta_{\text{ref}}}$ is the policy of a frozen reference model (usually the model state before unlearning), $\beta$ is a temperature parameter controlling the strength of the preference, and $\sigma$ is the sigmoid function.

We also consider the utility-keeping method as following:
\paragraph{Gradient Descent (GD)} applies the standard training objective on a curated retain set $S_r$, ensuring that the targeted unlearning process does not degrade overall utility:

\begin{equation*}
    \theta \leftarrow \theta - \eta \nabla_{\theta} \mathcal{L}_{\text{LM}}(S_r)
\end{equation*}

Here, $S_r$ is the retain set and $\theta$ represents the model's parameters.

\paragraph{Kullback–Leibler Divergence (KL)} acts as a regularization constraint to maintain model utility. It penalizes the deviation of the updated model's policy $\pi_{\theta}$ from that of a static reference model $\pi_{\theta_{\text{ref}}}$ on the retain set:

\begin{equation*}
    \mathcal{L}_{\text{KL}} = \mathbb{E}_{x \sim S_r} \left[ D_{\text{KL}}(\pi_{\theta_{\text{ref}}}(y|x) || \pi_{\theta}(y|x)) \right]
\end{equation*}

Here, $D_{\text{KL}}$ denotes the Kullback–Leibler divergence, which measures the difference between the probability distributions of the reference policy $\pi_{\theta_{\text{ref}}}$ and the current model policy $\pi_{\theta}$ over the retain set $S_r$.

Each unleanring pipelines can be paired with utility-keeping method.
\begin{table*}[htbp]
\centering
\scriptsize
\begin{tabular}{ll|cccc|cccc}
\toprule
Baseline & Generation Method
& \multicolumn{4}{c|}{Forget}
& \multicolumn{4}{c}{Utility} \\
&
& Prob$\downarrow$ & ROUGE$\downarrow$ & TruthRatio$\uparrow$ & Forget Q.$\uparrow$
& RS Score$\uparrow$ & RAS Score$\uparrow$ & WFS Score$\uparrow$ & Model Utility$\uparrow$ \\
\midrule
\multicolumn{2}{l|}{Before}
& 0.9951 & 0.9493 & 0.5306 & 0.0013
& 0.7133 & 0.1667 & 0.0508 & 0.6277 \\
\midrule
\multirow{3}{*}{GA+GD}
& TOFU
& \textbf{0.1031} & \textbf{0.1631} & \textbf{0.7365} & \textbf{0.2657}
& 0.1834 & \textbf{0.5251} & \textbf{0.3904} & 0.3497 \\
& DirectQA
& 0.9843 & 0.9198 & \underline{0.5395} & 0.0068
& \textbf{0.7015} & \underline{0.1852} & \underline{0.0616} & \textbf{0.6333} \\
\rowcolor{gray!15}
& \ourname\
& \underline{0.1608} & \underline{0.4416} & 0.5283 & \underline{0.1650}
& \underline{0.4748} & 0.0143 & 0.0048 & \underline{0.5436} \\
\midrule
\multirow{3}{*}{GA+KL}
& TOFU
& 0.4709 & 0.4409 & \textbf{0.5644} & \underline{0.0541}
& \textbf{0.5923} & \textbf{0.1617} & \textbf{0.0516} & \textbf{0.5943} \\
& DirectQA
& \underline{0.2916} & \underline{0.4092} & \underline{0.4950} & \underline{0.0541}
& \underline{0.5069} & \underline{0.0920} & \underline{0.0272} & \underline{0.5688} \\
\rowcolor{gray!15}
& \ourname\
& \textbf{0.1580} & \textbf{0.3359} & 0.4716 & \textbf{0.0971}
& 0.4767 & 0.0057 & 0.0028 & 0.5362 \\
\bottomrule
\end{tabular}
\caption{Unlearning performance(GA+GD, GA+KL) on TOFU-finetuned Llama-2-7b-chat. (↓ lower is better; ↑ higher is better)}
\label{tab:apx_tofu_7B}
\end{table*}

\begin{table*}[htbp]
\centering
\scriptsize
\begin{tabular}{ll|cccc|cccc}
\toprule
Baseline & Generation Method
& \multicolumn{4}{c|}{Forget}
& \multicolumn{4}{c}{Utility} \\
&
& Prob$\downarrow$ & ROUGE$\downarrow$ & TruthRatio$\uparrow$ & Forget Q.$\uparrow$
& RS Score$\uparrow$ & RAS Score$\uparrow$ & WFS Score$\uparrow$ & Model Utility$\uparrow$ \\
\midrule
\multicolumn{2}{l|}{Before}
& 0.7742 & 0.6469 & \textbf{0.5077} & 0.0068
& 0.5664 & \textbf{0.2048} & 0.1236 & 0.5735 \\
\midrule
\multirow{3}{*}{GA}
& TOFU
& \underline{0.2995} & \textbf{0.3632} & \textbf{0.6180} & \textbf{0.2657}
& 0.4173 & \textbf{0.5867} & \textbf{0.5741} & \underline{0.5222} \\
& DirectQA
& 0.5684 & \underline{0.4337} & \underline{0.5530} & 0.0143
& \textbf{0.5033} & \underline{0.2418} & \underline{0.1948} & \textbf{0.5586} \\
\rowcolor{gray!15}
& \ourname\
& \textbf{0.2799} & 0.5457 & 0.4635 & \underline{0.0286}
& \underline{0.4531} & 0.0059 & 0.0028 & 0.4948 \\
\midrule
\multirow{3}{*}{NPO}
& TOFU
& \textbf{0.2197} & \textbf{0.3577} & \textbf{0.6328} & \textbf{0.0971}
& 0.3290 & \underline{0.1771} & \underline{0.0832} & 0.4371 \\
& DirectQA
& 0.7731 & 0.6452 & \underline{0.5094} & 0.0068
& \textbf{0.5649} & \textbf{0.2088} & \textbf{0.1271} & \textbf{0.5733} \\
\rowcolor{gray!15}
& \ourname\
& \underline{0.4493} & \underline{0.6323} & 0.5033 & \underline{0.0286}
& \underline{0.5365} & 0.0239 & 0.0097 & \underline{0.5441} \\
\midrule
\multirow{3}{*}{GA+GD}
& TOFU
& \textbf{0.2912} & \textbf{0.3154} & \textbf{0.7063} & \textbf{0.2657}
& 0.3342 & \textbf{0.5469} & \textbf{0.4962} & 0.4592 \\
& DirectQA
& 0.5588 & \underline{0.4149} & \underline{0.5750} & 0.0068
& \underline{0.4463} & \underline{0.4221} & \underline{0.3572} & \textbf{0.5230} \\
\rowcolor{gray!15}
& \ourname\
& \underline{0.3165} & 0.5495 & 0.5030 & \underline{0.1650}
& \textbf{0.4894} & 0.0096 & 0.0043 & \underline{0.5096} \\
\midrule
\multirow{3}{*}{GA+KL}
& TOFU
& \textbf{0.1674} & \textbf{0.2680} & 0.4734 & \textbf{0.4046}
& 0.4772 & \underline{0.1876} & \underline{0.1108} & \underline{0.5453} \\
& DirectQA
& 0.6929 & 0.5873 & \textbf{0.5138} & 0.0068
& \textbf{0.5576} & \textbf{0.2068} & \textbf{0.1357} & \textbf{0.5749} \\
\rowcolor{gray!15}
& \ourname\
& \underline{0.5174} & \underline{0.5727} & \underline{0.4741} & \underline{0.0541}
& \underline{0.5467} & 0.0227 & 0.0121 & 0.5449 \\
\bottomrule
\end{tabular}
\caption{Unlearning performance with different forget set generation methods on TOFU using a TOFU-finetuned Llama-3.2-1B-Instruct model. (↓ lower is better; ↑ higher is better)}
\label{tab:tofu_small_llama3p2_1b}
\end{table*}

\begin{table*}[t]
\centering
\scriptsize
\setlength{\tabcolsep}{2.8pt}
\renewcommand{\arraystretch}{1.06}
\begin{tabular}{>{\centering\arraybackslash}m{1.45cm} c p{2.6cm} p{3.3cm} c c p{3.3cm} c c}
\toprule
\multirow{2}{*}{\textbf{Attributes}} & \multirow{2}{*}{\textbf{Freq}} & \multirow{2}{*}{\textbf{Gold}} &
\multicolumn{3}{c}{\textbf{DirectQA}} & \multicolumn{3}{c}{\textbf{\ourname}} \\
\cmidrule(lr){4-6}\cmidrule(lr){7-9}
& & & \textbf{Pred} & \textbf{F1} & \textbf{Covered} & \textbf{Pred} & \textbf{F1} & \textbf{Covered} \\
\midrule

BirthPlace & 2 & kuwait city, kuwait
& kuwait city, kuwait; & 1.00 & \ding{51}
& kuwait city; & 0.80 & \ding{51} \\
\midrule

Gender & 1 & male
&  & 0 & 
&  & 0 &  \\
\midrule

BirthYear & 1 & 1956
&  & 0 & 
& \makecell[l]{1931;\\1956;} & 0.67 & \ding{51} \\
\midrule

Genre & 3 & french literature
& \makecell[l]{Arabic culture and \\Middle Eastern landscapes;} & 0 & \ding{51}
& \makecell[l]{french literature;\\literary fiction;} & 0.67 & \ding{51} \\
\midrule

Awards & 1 & prix goncourt
&  & 0 & 
& prix goncourt; & 1.00 & \ding{51} \\
\midrule

Par\_Job & 4 & \makecell[l]{florist;\\game developer;}
& \makecell[l]{florist;\\game developer;\\accountant;} & 0.80 & \ding{51}
&  & 0 &  \\
\midrule

Books & 3 & \makecell[l]{promise by the seine;\\le petit sultan;}
&  & 0 & 
& \makecell[l]{promise by the seine;\\le petit sultan;\\the parisian promise;} & 0.80 & \ding{51} \\
\midrule

Books & 1 & \makecell[l]{river of jasmine;\\the nightingale of paris;}
&  & 0 & 
& \makecell[l]{the oasis of paris;\\mirror of paris;\\the nightingales song;\\the nightingale of paris;} & 0.33 & \ding{51} \\
\bottomrule
\end{tabular}
\caption{Qualitative comparison of recovered attributes against TOFU ground-truth profiles.}
\label{tab:memory_tofu_case}
\end{table*}

\section{Prompt Template} \label{apx:prompt}

\subsection{Prompt for internal memory mining}\label{apx:prompt_mining}
Prompt template for $E_t$'s 0-hop neighbor information mining.
\begin{mdframed}[style=promptbox]
\texttt{Recall information from your own internal knowledge.\\
Target: \{target\}\\
Write 5-10 atomic statements about \{target\}.}
\end{mdframed}

Prompt template for $h$-hop neighbor information mining, given the $h-1$-hop neighbor entity and the unlearning target.
\begin{mdframed}[style=promptbox]
\texttt{Recall information from your own internal knowledge.\\
Target: \{target\}\\
Neighbor: \{neighbor\}\\
Write 5-10 SHORT atomic statements specifically about how \{neighbor\} relates to \{target\}.
}
\end{mdframed}

\subsection{Prompt for corpus construction}\label{apx:prompt_corpus}

Prompt template for event statement construction, given the two neighbor entities from sampled path and the unlearning target.
\begin{mdframed}[style=promptbox]
\texttt{Please act as an information assistant to help users learn about pertinent details regarding the target.\\
Given an anchor target, according to known knowledge about the target, and two key event about the target, please provide ONE concise factual statement about the target's main information.\\
The statement should highlight details about the \{target\} that users may find important.\\
Do NOT invent fictional or hypothetical scenarios. If you are not confident the connection is real, output UNKNOWN.\\
The statement should explicitly include BOTH eventss\' names (do not use pronouns), and keep it to ONE sentence.\\
Anchor Target: \{target\}\\
Event 1: \{event\_1\}\\
Event 2: \{event\_2\}
}
\end{mdframed}

Prompt template for QA-style forget set construction, given the event statement, context obj and the unlearning target.

\begin{mdframed}[style=promptbox]
\texttt{Please act as an information assistant to help users learn about pertinent details.\\
Given a factual statement about the target, rewrite it into ONE QA pair.\\
The question should highlight an important detail from the statement about Contral Context.\\
The question MUST include the target entity's name (do not use pronouns).\\
Target Entity: \{target\}\\
Central Context (Obj): \{obj\}\\
Statement: \{event\}\\
}
\end{mdframed}

\section{More Results}

\subsection{Unlearning Results in Large Model}\label{apx:large_res_tofu}
Table~\ref{tab:apx_tofu_7B} details the results on Llama-2-7b-chat. Under the \textbf{GA+GD} setting, \ourname\ achieves a superior balance compared to baselines: it avoids the unlearning failure seen in DirectQA while maintaining significantly higher utility than the TOFU supervision, which suffers from severe catastrophic forgetting. Notably, under the \textbf{GA+KL} constraint, \ourname\ demonstrates the strongest forgetting capabilities among all methods, achieving the lowest Prob and ROUGE scores. This confirms that our mined memory graph provides precise supervision even under strict regularization.

\subsection{Unlearning Results in Smaller Model}\label{apx:small_res}
\subsubsection{RWKU Evaluation on Phi-3-mini-4k}\label{apx:res_phi3}
Table~\ref{tab:apx_rwku_res} shows results on Phi-3-mini-4k. \ourname\ demonstrates robust forgetting across objectives, notably achieving the lowest aggregate Forget Set score in GA and significantly outperforming baselines in NPO. Crucially, it offers a superior utility-forgetting trade-off: unlike ELUDe which suffers severe reasoning degradation (e.g., sharp BBH drop in GA+GD), \ourname\ maintains high performance across utility metrics, validating its precision in targeting specific memories without compromising general capabilities.

\subsubsection{TOFU Evaluation on Llama-3.2-1B}\label{apx:res_llama3_1b}
Table~\ref{tab:tofu_small_llama3p2_1b} presents the unlearning results on Llama-3.2-1B-Instruct. \ourname\ exhibits a distinct advantage over the DirectQA baseline, which struggles to induce effective forgetting in this smaller architecture (e.g., yielding retention scores comparable to the pre-unlearning state under NPO and GA+GD). In contrast, \ourname\ consistently reduces target knowledge retention by leveraging mined internal memory signals, providing the necessary supervision that standard prompting lacks, while even outperforming the gold-standard TOFU supervision in specific settings (e.g., GA).

\section{Additional Analysis}
\subsection{TOFU Recovery Details}\label{apx:tofu_memory_detail}
We provide detailed TOFU attribute recovery results using ``Basil Mahfouz Al-Kuwaiti'' as an illustrative example in Table~\ref{tab:memory_tofu_case}.

\begin{figure}[t]
    \centering
    \includegraphics[width=0.75\linewidth]{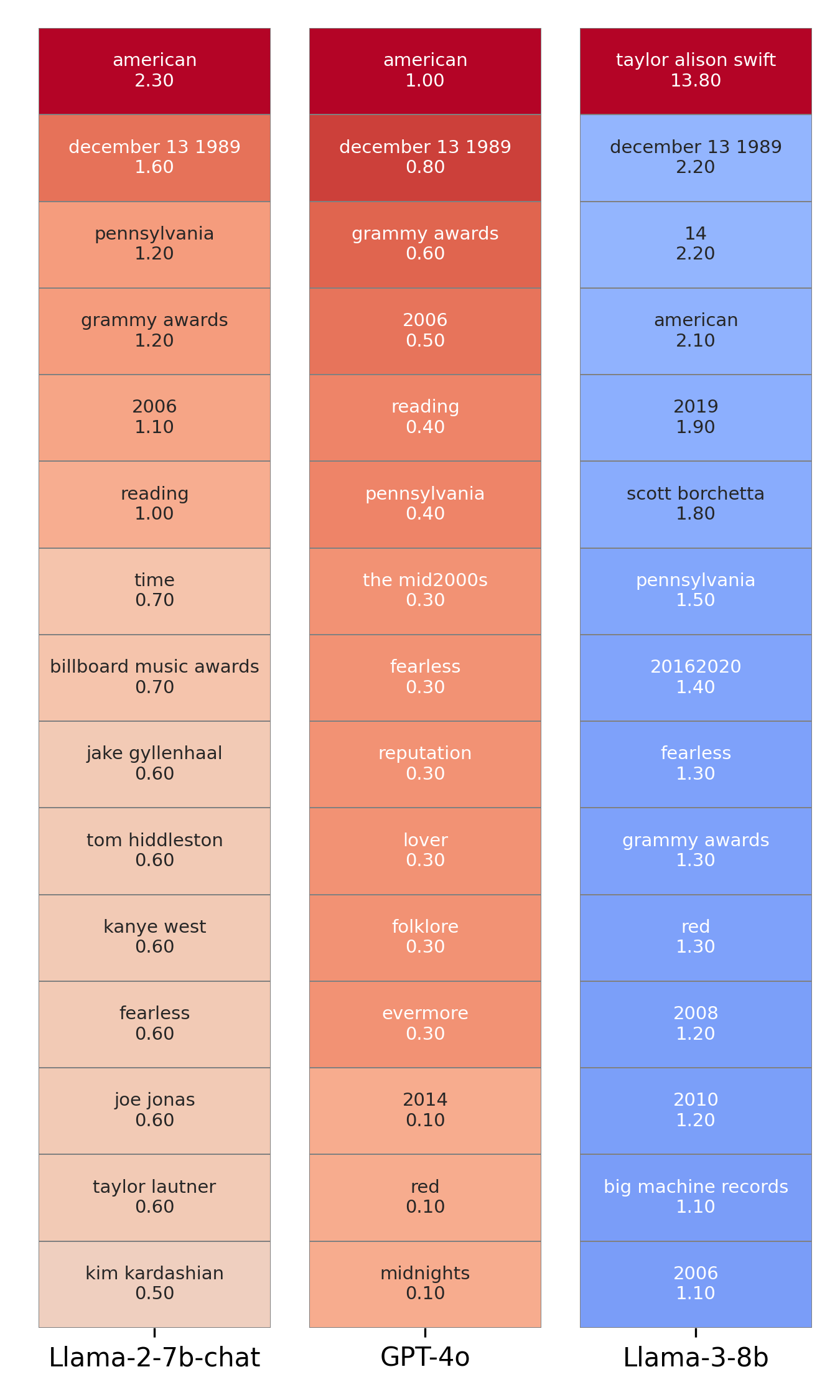}
    \caption{Memory strength variation of 1-hop neighbors across models.}
    \label{fig:heatmap}
\end{figure}
\subsection{Variation Across Models} \label{apx:memory_variation}

We observe substantial cross-model variation in recovered memory graphs, even for the same entity. For “Taylor Swift” in RWKU, graphs from LLaMA-2-7b-chat, LLaMA-3-8b-instruct, and GPT-4o exhibit low node overlap and distinct memorization patterns (Figure~\ref{fig:heatmap}). While all three models share a few core biographical anchors—most notably “december 13 1989” and “american”, their relative strengths differ markedly.
Notably, LLaMA-3-8b consistently recalls “Taylor Alison Swift” as a salient neighbor, which is absent in the others.
These differences, likely driven by training data and architectural choices, suggest that forget supervision should be regenerated per target model to ensure effective unlearning.

\subsection{Knowledge Coverage} \label{apx:knowledge_coverage}
\begin{figure}[t]
    \centering
    \includegraphics[width=\linewidth]{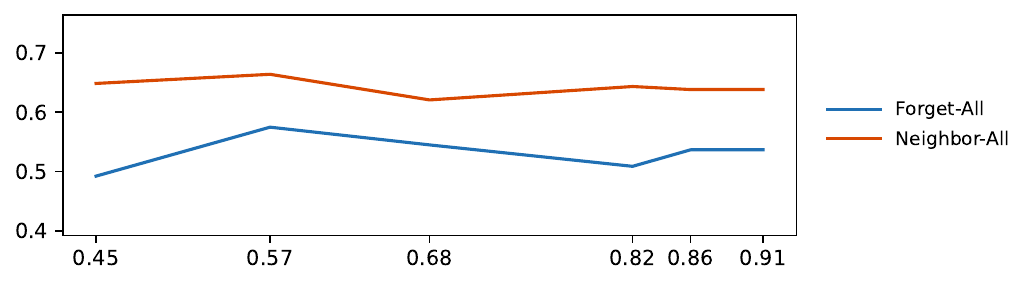}
\caption{Impact of Memory Graph Coverage on Unlearning Performance.}
    \label{fig:coverage_ratio}
\end{figure}

Specifically, on RWKU with \texttt{Llama-2-7b-chat}, we vary the coverage threshold used during forget-set construction from the recovered memory graph. For each threshold, we compute the average coverage of the resulting forget set over the memory graph across all target entities, and report the corresponding unlearning results in Figure~\ref{fig:coverage_ratio}.

\end{document}